\documentclass[final]{cvpr}

\usepackage{times}
\usepackage{epsfig}
\usepackage{graphicx}
\usepackage{amsmath}
\usepackage{amssymb}
\usepackage{enumitem}
\usepackage{booktabs}

\usepackage{slashbox}
\usepackage{subcaption}
\usepackage{pifont}
\newcommand{\cmark}{\ding{52}}
\newcommand{\xmark}{\ding{56}}
\usepackage{multirow}
\usepackage{makecell}
\usepackage{boldline}
\setcellgapes{3pt}
\usepackage[symbol]{footmisc}

\usepackage[pagebackref=true,breaklinks=true,colorlinks,bookmarks=false]{hyperref}

\def\cvprPaperID{4697} 
\def\confYear{CVPR 2021}

\begin{document}

\title{Look Closer to Segment Better: \\ Boundary Patch Refinement for Instance Segmentation}

\author{Chufeng Tang$^{1}$\thanks{Equal contribution.} \ \ Hang Chen$^{1}$\footnotemark[1] \ \ Xiao Li$^{1}$ \ \ Jianmin Li$^{1}$ \ \ Zhaoxiang Zhang$^{2}$ \ \ Xiaolin Hu$^{1}$\thanks{Corresponding author.} \\
$^{1}$State Key Laboratory of Intelligent Technology and Systems, THU-Bosch JCML Center, \\ BNRist, Institute for AI, Department of Computer Science and Technology, Tsinghua University \\
$^{2}$Institute of Automation, CAS \& University of Chinese Academy of Sciences \& \\ Centre for Artificial Intelligence and Robotics, HKISI\_CAS \\
{\tt\small \{tcf18, chenhang20, lixiao20\}@mails.tsinghua.edu.cn \ zhaoxiang.zhang@ia.ac.cn} \\ \tt\small \{lijianmin, xlhu\}@mail.tsinghua.edu.cn}

\maketitle
\pagestyle{empty}
\thispagestyle{empty}

\begin{abstract}
  Tremendous efforts have been made on instance segmentation but the mask quality is still not satisfactory.
  The boundaries of predicted instance masks are usually imprecise due to the low spatial resolution of feature maps and the imbalance problem caused by the extremely low proportion of boundary pixels.
  To address these issues, we propose a conceptually simple yet effective post-processing refinement framework to improve the boundary quality based on the results of any instance segmentation model, termed BPR.
  Following the idea of looking closer to segment boundaries better, we extract and refine a series of small boundary patches along the predicted instance boundaries.
  The refinement is accomplished by a boundary patch refinement network at higher resolution.
  The proposed BPR framework yields significant improvements over the Mask R-CNN baseline on Cityscapes benchmark, especially on the boundary-aware metrics.
  Moreover, by applying the BPR framework to the ``PolyTransform + SegFix'' baseline, we reached $1^{st}$ place on the Cityscapes leaderboard.
  Code is available at \url{https://github.com/tinyalpha/BPR}.
\end{abstract}

\section{Introduction} \label{sec:introduction}
Instance segmentation, which aims to assign a pixel-wise instance mask with a category label to each object in an image, has great potential in various computer vision applications, such as autonomous driving and robotics.
Mask R-CNN \cite{maskrcnn} is a prevailing two-stage instance segmentation framework, which first employs a Faster R-CNN \cite{fastrcnn} detector to detect objects in an image and further performs binary segmentation within each detected bounding box.
Other methods \cite{msrcnn,panet} built upon Mask R-CNN consistently achieve superior performance.
Driven by the recent development of one-stage detectors \cite{retinanet,fcos,centernet}, a number of one-stage instance segmentation frameworks \cite{yolact,blendmask,chen2019tensormask, lee2020centermask,condinst,solo,solov2,xie2020polarmask,ying2019embedmask,MEInst} have been proposed.

\begin{figure}[t]
\begin{center}
  \includegraphics[width=1.0\linewidth]{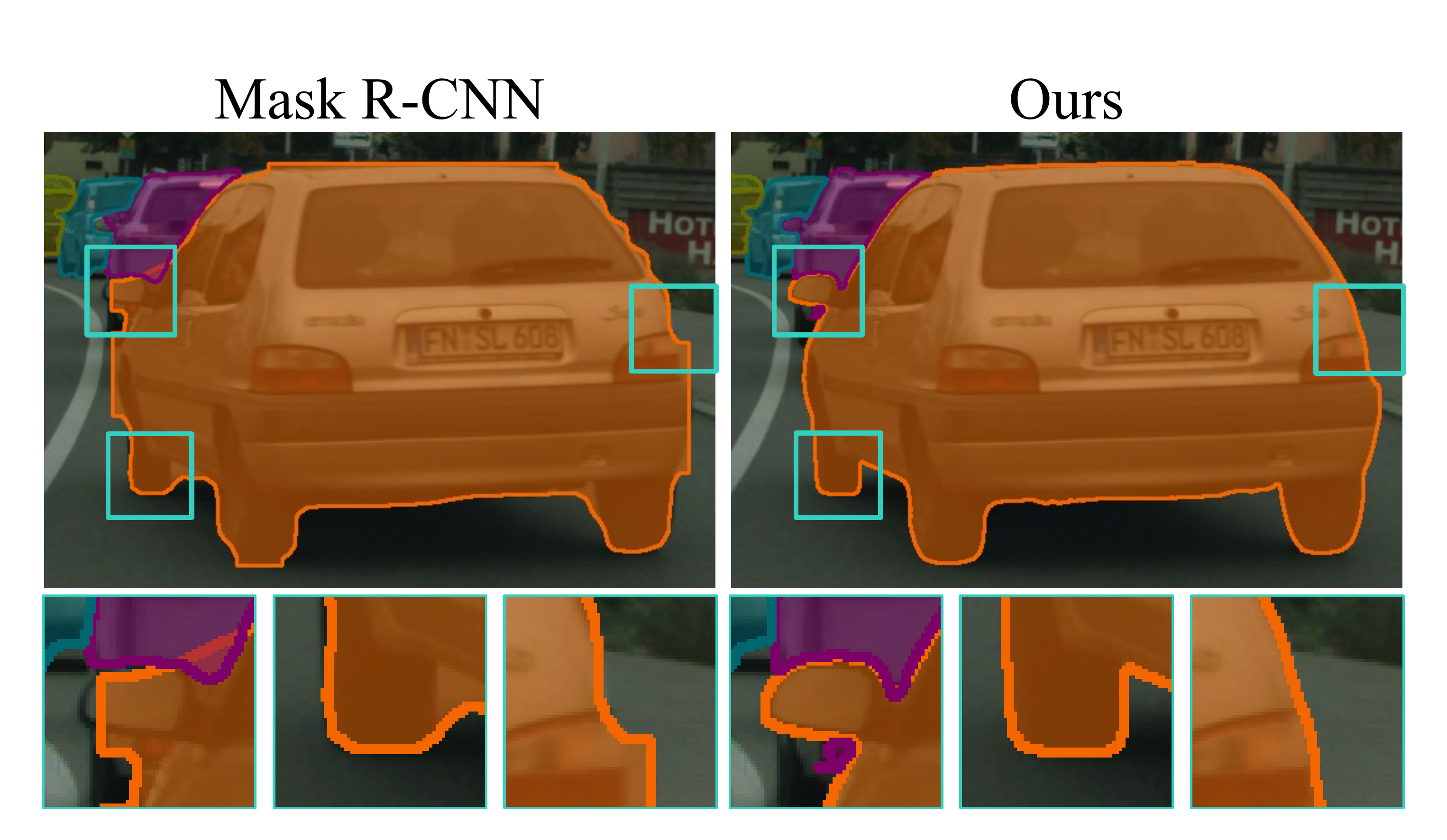}
\end{center}
  \vspace{-6mm}
   \caption{
     \textbf{Left}: Instance segmentation results and the extracted boundary patches of Mask R-CNN.
     \textbf{Right}: After the refinement of our BPR framework, the instance mask aligns better with object boundaries.
     Best viewed with zoom-in.
   }
   \label{fig:introduction}
\end{figure}

However, the quality of the predicted instance mask is still not satisfactory.
One of the most important problems is the imprecise segmentation around instance boundaries.
As shown in Figure~\ref{fig:introduction}(left), the predicted instance masks of Mask R-CNN are coarse and not well-aligned with the real object boundaries.
Empirically, correcting the error pixels near object boundaries can improve the mask quality a lot.
We conducted an upper bound analysis in Table~\ref{Tab:intro_gt_improves}.
A large gain (9.4/14.2/17.8 in AP) can be obtained by simply replacing the predictions with ground-truth labels for pixels within a certain Euclidean distance (1px/2px/3px) to the predicted boundaries, especially for small objects.

We argue that there are two critical issues leading to low-quality boundary segmentation.
(1) The low spatial resolution of the output, \eg 28$\times$28 in Mask R-CNN or at most 1/4 input resolution in some one-stage frameworks \cite{condinst,solov2}, makes finer details around object boundaries disappear.
The predicted boundaries are always coarse and imprecise (see Figure~\ref{fig:introduction} and \ref{fig:visualization}).
(2) Pixels around object boundaries only make up a small fraction of the whole image (less than 1\% \cite{kirillov2017instancecut}), and are inherently hard to classify.
Treating all pixels equally may leads to an optimization bias towards smooth interior areas, while underestimating the boundary pixels.

As a long-standing challenge in dense prediction tasks, many studies have attempted to improve the boundary quality, while the above issues are still not well solved.
For example, BMask R-CNN \cite{bmaskrcnn} and Gated-SCNN \cite{gatedscnn} employ an extra branch to enhance the boundary awareness of mask features, which can fix the optimization bias to some extent, while the low resolution issue remains unsolved.
PolyTransform \cite{polytransform} and SegFix \cite{segfix} act as a post-processing scheme to improve the boundary quality.
PolyTransform \cite{polytransform} employs a deforming network with the cropped instance patch to predict the offsets of polygon vertices, while suffering from a large computational overhead.
SegFix \cite{segfix} replaces the coarse predictions of boundary pixels with interior predictions, but it relies on precise boundary predictions.
We argue that the instance boundary prediction task shares a similar complexity with instance segmentation.

Considering the human annotation behavior for instance segmentation, the annotators usually first localize and categorize each object in the given image, and then explicitly or implicitly segment some coarse instance masks at a low resolution.
Afterwards, to obtain a high-quality mask, the annotators need to repeatedly zoom into the local boundary regions and explore the sharper boundary segmentation at higher resolution.
Intuitively, high-level semantics are required to localize and roughly segment objects, while low-level details (\eg colour consistency and contrast) are more critical for segmenting the local boundary regions.

In this paper, motivated by the human segmentation behavior, we propose a conceptually simple yet effective post-processing framework to improve the boundary quality through a \textit{crop-then-refine} strategy.
Specifically, given a coarse instance mask produced by any instance segmentation model, we first extract a series of small image patches along the predicted instance boundaries.
After concatenated with mask patches, the boundary patches are fed into a refinement network, which performs binary segmentation to refine the coarse boundaries.
The refined mask patches are then reassembled into a compact and high-quality instance mask, shown in Figure~\ref{fig:introduction}(right).
We termed the proposed framework as \textbf{BPR} (\textbf{B}oundary \textbf{P}atch \textbf{R}efinement).
The proposed framework can alleviate the aforementioned issues and improve the mask quality without any modification or fine-tuning to the segmentation models.
Since we only crop around object boundaries, the patches are allowed to be processed with the much higher resolution than previous methods, so that low-level details can be retained better. 
Concurrently, the fraction of boundary pixels in the small patch is naturally increased, which can alleviate the optimization bias.
The proposed BPR framework significantly improves the results of Mask R-CNN baseline (+$4.3\%$ AP on Cityscapes dataset), and produces substantially better masks with finer boundaries.
We found that the model trained on the results of Mask R-CNN can be easily transferred to refine the results of other instance segmentation models as well, without the need for re-training.
We outperform some boundary refinement methods \cite{pointrend,segfix} and show that these methods are complementary by successfully transferring our model to improve their results.
Furthermore, by applying our BPR framework to the ``PolyTransform + SegFix'' baseline \cite{segfix}, we established a new state-of-the-art on the Cityscapes test set with AP of $42.7\%$, and ranked $1^{st}$ place
on the Cityscapes leaderboard by the CVPR 2021 submission deadline.

\begin{table}[t]
\begin{center}
  \begin{tabular}{c|ccc|ccc}
    Dist. &AP &AP$_{50}$ &AP$_{75}$ &AP$_{S}$ &AP$_{M}$ &AP$_{L}$  \\ \hline
    -   &36.4 &60.8 &36.9 &11.1 &32.4 &57.3 \\ \hline
    1px &45.8 &64.8 &49.3 &21.1 &42.6 &63.5 \\
    2px &50.6 &66.5 &54.6 &26.3 &47.0 &66.8 \\
    3px &54.2 &67.5 &58.5 &30.4 &50.7 &69.3 \\
    $\infty$ &70.4 &70.4 &70.4 &41.5 &66.7 &88.3 \\
  \end{tabular}
\end{center}
\vspace{-5mm}
  \caption{
    A large gain can be obtained by replacing the predictions for pixels within a certain Euclidean distance to the predicted boundaries with their group-truth labels.
    $\infty$ means all error pixels are corrected.
    Experiments were conducted with Mask R-CNN as baseline on Cityscapes val set.
  }
  \label{Tab:intro_gt_improves}
\end{table}

\section{Related Work} \label{sec:related-works}

\textbf{Instance Segmentation}.
Recent studies on instance segmentation can be divided into two categories: two-stage and one-stage methods, as briefly reviewed below.

Two-stage methods usually follow the classical \textit{detect-then-segment} strategy.
The dominant method is still Mask R-CNN \cite{maskrcnn}, which inherits from the two-stage detector Faster R-CNN \cite{fastrcnn} to first detect objects in an image and further performs binary segmentation within each detected bounding box.
Following Mask R-CNN, PANet \cite{panet} enhances feature representation through bottom-up path augmentation.
Mask Scoring R-CNN \cite{msrcnn} adds an additional mask-IoU head to re-score the mask predictions.
These methods consistently achieve superior performance.

One-stage methods recently attracts more attention due to the rapid development of one-stage detectors \cite{retinanet,fcos,centernet}.
Some methods \cite{yolact,blendmask,lee2020centermask,ying2019embedmask,MEInst} continue to adapt the \textit{detect-then-segment} strategy but replace the detectors with the one-stage alternatives.
YOLACT \cite{yolact} achieves real-time speed by learning a set of prototypes and the prototypes are assembled with the learned linear coefficients.
BlendMask \cite{blendmask} further improves this idea by assembling with attention maps.
Some recent proposed methods \cite{chen2019tensormask,condinst,solo,solov2} eliminate the need for detection by directly segmenting objects in a location-wise manner.
CondInst \cite{condinst} and SOLOv2 \cite{solov2} achieve remarkable performance with high efficiency.
In addition, there are some approaches \cite{de2017semantic,SSAP,neven2019instance} built upon the semantic segmentation models, which usually learn the pixel-wise embeddings and then cluster them into instances.
Several works \cite{DWT,deepsnake,xie2020polarmask,xu2019explicit} replace the pixel-wise instance representation into the contour-based representation.

Our proposed framework is agnostic to the instance segmentation methods, thus it can be applied to refine the results of any instance segmentation model, both one-stage and two-stage methods.

\textbf{Semantic Segmentation}.
Modern semantic segmentation approaches are pioneered by fully convolutional networks (FCNs) \cite{fcn}.
Many studies have been proposed on this foundation to improve the segmentation results,
such as increasing the resolution of feature maps with dilated/atrous convolutions \cite{chen2017rethinking,chen2018encoder},
enriching context information \cite{fu2019dual,ocrnet,zhang2019co,zhao2017pyramid},
using an encoder-decoder architecture \cite{chen2018encoder,kirillov2019panoptic,milletari2016v,ronneberger2015u},
or some refinement schemes \cite{krahenbuhl2011efficient,li2016iterative,segfix}.
Minaee \etal \cite{minaee2020image} provided a comprehensive review of these approaches.
In this paper, we adopt the prevailing HRNet \cite{hrnet} in our framework, which can maintain high-resolution representation throughout the whole network.

\textbf{Boundary Refinement for Segmentation}.
Most recent studies focused on boundary refinement aim at designing a boundary-aware segmentation model by integrating an extra and specialized module to process boundaries.
For example, BMask R-CNN \cite{bmaskrcnn} and Gated-SCNN \cite{gatedscnn} employ an extra branch to enhance the boundary awareness of mask features by estimating boundaries directly.
PointRend \cite{pointrend} iteratively samples the feature points with unreliable predictions and refines them with a shared MLP.
Another line of work attempts to refine the boundaries based on the results of existing segmentation models with a post-processing scheme.
SegFix \cite{segfix} is a general refinement mechanism, which replaces the unreliable predictions of boundary pixels with the predictions of interior pixels.
The effectiveness of SegFix highly depends on the accuracy of boundary prediction.
However, it is very challenging to directly estimate precise instance boundaries.
Intuitively, the instance segmentation task could easily be settled if the precise boundaries are already given.
Our method shares more similarities with PolyTransform \cite{polytransform}, which transforms the contour of instance into a set of polygon vertices.
A Transformer \cite{vaswani2017attention} based network is applied to predict the offsets of vertices towards object boundaries.
It achieves superior performance while suffering from a large computational overhead due to the large instance patch and the heavy Transformer architecture.
Our proposed method is also a post-processing scheme. Different from these methods, we focus on refining the boundary patches to improve the mask quality.

\begin{figure*}[t]
\begin{center}
  \includegraphics[width=0.84\linewidth]{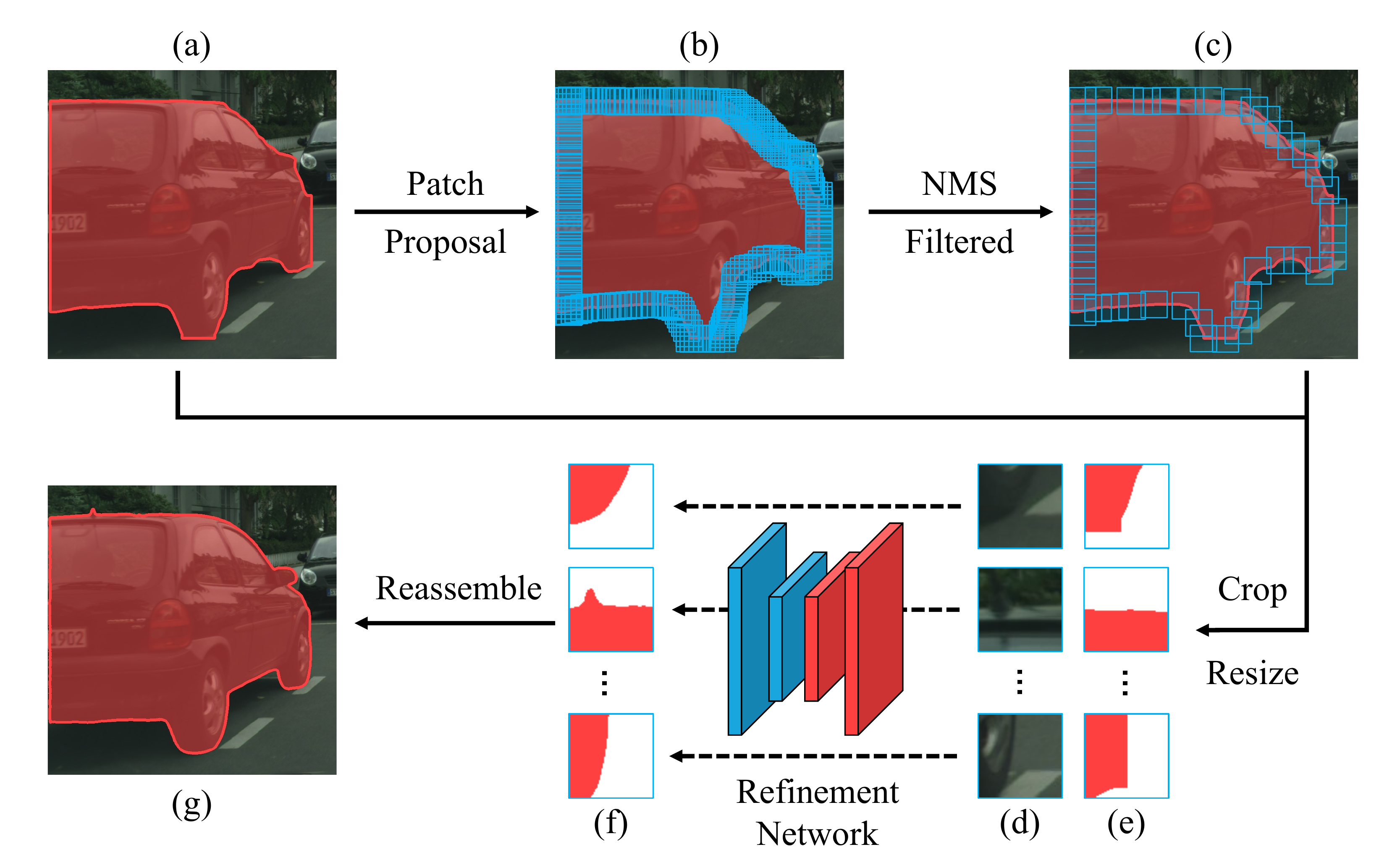}
\end{center}
\vspace{-6mm}
  \caption{
    Overview of the proposed boundary patch refinement framework.
    Given a coarse mask (a) produced by an instance segmentation model, we first densely assign a series of squared bounding boxes along the predicted boundaries (b),
    and filter out a subset of boundary patches (c) using NMS. NMS threshold of 0.25 is used here.
    The extracted image patches (d) and mask patches (e) are resized and fed into the boundary patch refinement network.
    Mask patches after refinement (f) are reassembled into a compact and precise instance mask (g). Best viewed digitally and in colour.
  }
  \label{fig:framework}
\end{figure*}

\section{Framework} \label{sec:framework}
An overview of the proposed framework is illustrated in Figure~\ref{fig:framework}.
As a post-processing mechanism, the proposed framework can be applied to refine the results of any prevailing instance segmentation model, without any modification or fine-tuning to the segmentation models themselves.

\subsection{Boundary Patch Extraction}
Given an instance mask produced by an instance segmentation model, we first need to determine which part of the mask should be refined.
Based on the findings of previous works \cite{bmaskrcnn,segfix} and our verification experiments in Table~\ref{Tab:intro_gt_improves}, we propose an effective \textit{sliding-window} style algorithm to extract a series of patches along the predicted instance boundaries.
Specifically, we densely assign a group of squared bounding boxes where the central areas of the box should cover the boundary pixels, shown in Figure~\ref{fig:framework}(b).
The obtained boxes still contain large overlaps and redundancies, thus we further apply a Non-Maximum Suppression (NMS) algorithm to filter out a subset of patches (Figure~\ref{fig:framework}c).
Empirically, with the larger overlaps, the segmentation performance can be boosted, while simultaneously suffering from the larger computational cost.
We can adjust the NMS threshold to control the amount of overlap to achieve a better speed/accuracy trade-off.
In addition to image patches, we also extract the corresponding binary mask patches from the given instance mask.
The concatenated image and mask patches (Figures~\ref{fig:framework}d and \ref{fig:framework}e) are resized and fed into the following boundary patch refinement network.

\subsection{Boundary Patch Refinement}

\textbf{Mask Patch}.
The benefit of the binary mask patch is that it accelerates training convergence and provides location guidance for the instance to be segmented.
As discussed in the previous works on semantic segmentation \cite{hrnet,ocrnet}, context information plays a vital role for pixel-wise classification.
Therefore, the cropped image patches are hard to be classified independently due to the limited context information.
With the help of location and semantic information provided by the mask patches, the refinement network can eliminate the need for learning instance-level semantics from scratch.
Instead, the refinement network only needs to learn how to locate the hard pixels around the decision boundary and push them to the correct side.
We believe this goal can be achieved by exploring low-level image properties (\eg colour consistency and contrast) provided in the local and high-resolution image patches.
More importantly, the adjacent instances are likely to share an identical boundary patch, while the learning goals are totally different and ambiguous.
Together with different mask patches for each instance, these issues can be avoided.
As compared in Table~\ref{Tab:mask_patch_effects}, the model has trouble to converge without using the mask patches, examples of which are shown in Figure~\ref{fig:mask_patch}.

\textbf{Boundary Patch Refinement Network}.
The role of this refinement network is to perform binary segmentation for each extracted boundary patch individually.
Any semantic segmentation model can be employed for this task by simply modifying the input channels to 4 (3 for the RGB image patch and 1 for the binary mask patch) and output classes to 2.
For the sake of convenience, we adopt the state-of-the-art HRNetV2 \cite{hrnet} as the refinement network in our implementation, which can maintain high-resolution representation throughout the whole network.
By increasing the input size appropriately, the boundary patches can be processed with much higher resolution than in previous methods.

\textbf{Reassembling}.
The refined boundary patches are reassembled into a compact instance-level mask by replacing their previous predictions.
Predictions are unchanged for those pixels without refinement.
For the overlapping areas of adjacent patches, the results are aggregated by simply averaging the output logits and applying a threshold of 0.5 to distinguish the foreground and background.

\subsection{Learning and Inference}
The refinement network is trained based on the boundary patches extracted from training images and tested on validation or testing images.
We do not directly train or fine-tune the instance segmentation models.
During training, we only extract boundary patches from instances whose predicted masks have an Intersection over Union (IoU) overlap larger than 0.5 with the ground-truth masks, while all predicted instances are retained during inference.
The model outputs are supervised with the corresponding ground-truth mask patches using the pixel-wise binary cross-entropy loss.
We simply fix the NMS eliminating threshold to 0.25 during training, while adopting different thresholds during inference based on the speed requirements.
See Appendix \ref{appendix_a} for more implementation details.

\section{Experiments} \label{sec:experiments}

\subsection{Datasets and Metrics}

\textbf{Datasets}.
We mainly report the results on Cityscapes \cite{cityscapes}, a real-world dataset with high-quality instance segmentation annotations.
We only used the \texttt{fine} data, containing $2,975/500/1,525$ images for train/val/test, which are collected from 27 cities, with a high resolution of 1024$\times$2048.
Eight instance categories are involved, including bicycle, bus, person, train, truck, motorcycle, car, and rider.

\textbf{Metrics}.
The COCO-style \cite{mscoco} mask AP (averaged over 10 IoU thresholds ranging from 0.5 to 0.95 in the step of 0.05), AP$_{50}$ (AP at an IoU of 0.5) and AP$_S$/AP$_{M}$/AP$_{L}$ (for small/medium/large instances) were reported in most of our experiments.
The official Cityscapes-style AP \cite{cityscapes} was only used to report the final results for a fair comparison, which is slightly higher than the COCO-style AP.
Similar to \cite{polytransform,gatedscnn,segfix}, we also used a boundary F-score to evaluate the quality of the predicted boundaries.
A mask was considered correct if the boundary is within a certain distance threshold from the ground-truth.
We used a threshold of 1px and only compute for true positives that are determined on the same 10 IoU thresholds ranging from 0.5 to 0.95.
The boundary F-score was computed in an instance-wise manner and then averaged over them, termed AF.

\subsection{Ablation Study} \label{ablation_study}
We investigated the effectiveness of the proposed framework through extensive ablation experiments on the configurable design choices.
We started the refinement with the results of Mask R-CNN ResNet-FPN-50 baseline trained on Cityscapes \texttt{fine} data (with COCO pre-training).
We adopted the lightweight HRNetV2-W18-Small as the refinement network in the default setting, with input size equal to 128$\times$128.
The boundary patches were extracted with patch size equal to 64$\times$64 without padding, and the inference NMS threshold was set to 0.25 by default.

\begin{figure}[h]
\begin{center}
  \includegraphics[width=1.0\linewidth]{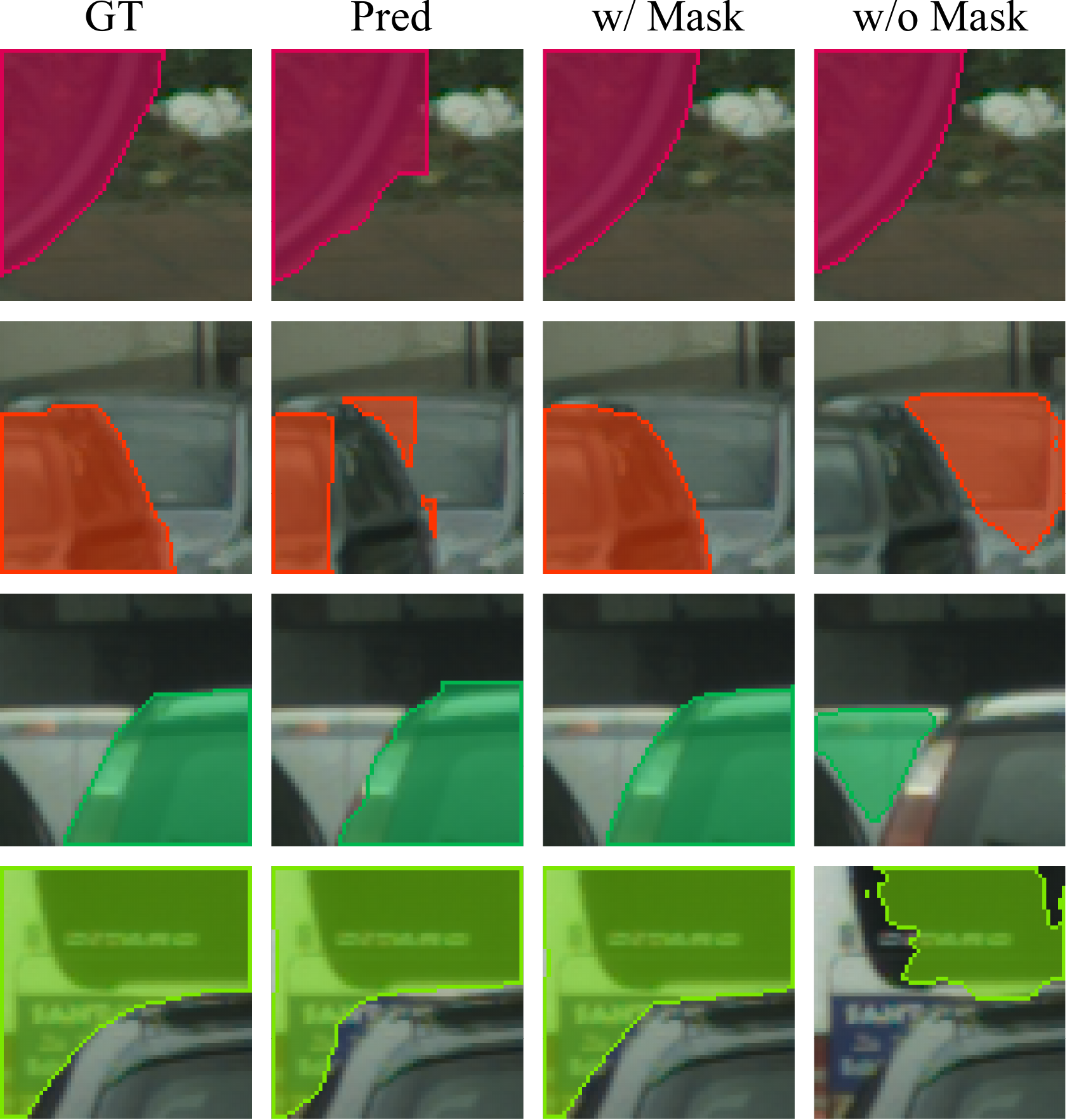}
\end{center}
\vspace{-5mm}
  \caption{
    Boundary patch examples of (from left to right): ground-truth, predictions of Mask R-CNN, results refined by our proposed framework, results without the use of mask patch.
    The mask patch plays a crucial role in our framework, resulting in high-quality boundaries (the 3rd column).
  }
  \label{fig:mask_patch}
\end{figure}

\textbf{Effects of Mask Patch}.
To validate the effect of mask patch for boundary refinement, we made a comparison by eliminating the mask patches while keeping other settings unchanged.
As indicated in Table~\ref{Tab:mask_patch_effects}, the model trained with image patches solely yielded a terrible result, even worse than the segmentation results before refinement.
However, together with mask patches, we achieved a significant improvement (+$3.4\%$ in AP, +$11.9\%$ in AF) by refining the Mask R-CNN segmentation results.
We also show some patch-wise examples in Figure~\ref{fig:mask_patch}.
For a simple case with one dominant instance in the image patch (first row), both of the models (w/ or w/o mask patch) produced reasonable results.
However, as for cases with multiple instances crowded in the image patch, the model without mask patch (last column) failed to distinguish which object should be segmented, leading to coarse (4th row) or completely wrong (2nd and 3rd rows) predictions.
In contrast, with the help of mask patches, we produced high-quality predictions with accurate and distinct boundaries (3rd column).

\begin{table}[t]
\begin{center}
  \begin{tabular}{c|cc|c|ccc}
    w/ mask  &AP &AP$_{50}$ &AF &AP$_{S}$ &AP$_{M}$ &AP$_{L}$ \\ \hline
      --      &36.4 &60.8 &54.9 &11.1 &32.4 &57.3 \\ \hline
    \xmark            &20.1 &42.2 &57.2 &4.0 &14.7 &36.3 \\
    \cmark  &\textbf{39.8} &\textbf{62.0} &\textbf{66.8} &\textbf{12.7} &\textbf{35.9} &\textbf{62.2} \\
  \end{tabular}
\end{center}
  \vspace{-5mm}
  \caption{
    \textbf{Effects of mask patch}: A dramatic performance drop can be observed without the use of mask patch.
    ``--'' indicates the results of Mask R-CNN before refinement.
  }
  \label{Tab:mask_patch_effects}
\end{table}

\textbf{Patch Size}.
We increased the boundary patch size by cropping with a larger box and/or with padding.
Note that the padded areas were only used to enrich the context and not used for reassembling.
As the patch size gets larger, the model becomes less focused but can access more context information.
In Table~\ref{Tab:patch_size}, we compared various choices and found that the 64$\times$64 patch without padding works better.
We used this setting in all experiments.

\begin{table}[t]
\begin{center}
  \begin{tabular}{c|cc|c|ccc}
    scale/pad  &AP &AP$_{50}$ &AF &AP$_{S}$ &AP$_{M}$ &AP$_{L}$ \\ \hline
       --     &36.4 &60.8 &54.9 &11.1 &32.4 &57.3 \\ \hline
     32 / 0   &39.4 &62.0 &66.8 &12.6 &35.6 &61.4 \\
     32 / 5   &39.7 &\textbf{62.2} &\textbf{67.6} &\textbf{12.9} &\textbf{35.9} &61.6 \\
     64 / 0   &\textbf{39.8} &62.0 &66.8 &12.7 &\textbf{35.9} &62.2 \\
     64 / 5   &39.7 &61.7 &66.5 &12.5 &35.8 &62.1 \\
     96 / 0   &39.6 &62.0 &65.7 &12.2 &35.4 &\textbf{62.3} \\
  \end{tabular}
\end{center}
\vspace{-5mm}
  \caption{
    \textbf{Results of different patch size}.
    The patch size of 64$\times$64 without padding works better.
  }
  \label{Tab:patch_size}
\end{table}

\textbf{Different Patch Extraction Schemes}.
The most important contribution of this work is the idea of looking closer at instance boundaries to achieve better segmentation results.
There are multiple choices about how to extract the boundary patches for refinement.
We compared three extraction schemes and listed the results in Table~\ref{Tab:diff_patch_extract}.
The most straightforward scheme is dividing the input image into a group of patches according to the pre-defined grids, and then picking only the patches that covering the predicted boundaries.
We varied the patch size and found the results were consistently worse than our proposed ``dense sampling + NMS filtering'' scheme.
One of the most important reasons is the imbalanced foreground/background ratio.
We observed that some extracted patches are almost entirely filled with either foreground or background pixels. These patches are hard to refine due to the lack of context, thus leading to sub-optimal results.
In contrast, by restricting the center of patches to cover the boundary pixels, the imbalance problem can be alleviated.
Another scheme, similar to some previous works \cite{polytransform,liu20201st}, is cropping the \textit{whole} instance based on the detected bounding box and further re-segmenting the instance patch.
As shown, even though the input size was increased to 512$\times$512, the results are still sub-optimal, which demonstrated the effectiveness of our \textit{local} boundary patches.
See Appendix \ref{appendix_b} for detailed descriptions.

\begin{table}[t]
\begin{center}
  \begin{tabular}{cc|cc|c}
      scheme &size &AP &AP$_{50}$ &AF \\ \hline
      --  &--  &36.4 &60.8 &54.9 \\ \hline
      dense sampling + NMS &64 &\textbf{39.8} &\textbf{62.0} &\textbf{66.8} \\ \hline
      pre-defined grid &32 &39.3 &61.8 &65.8 \\
      pre-defined grid &64 &39.1 &61.9 &65.6 \\
      pre-defined grid &96 &38.8 &61.6 &63.7 \\ \hline
      instance-level patch &256 &37.5 &61.1 &61.5 \\
      instance-level patch &512 &38.7 &61.6 &63.8 \\
  \end{tabular}
\end{center}
\vspace{-6mm}
  \caption{
    \textbf{Different patch extraction schemes}:
    The ``dense sampling + NMS filtering'' scheme works better.
  }
  \label{Tab:diff_patch_extract}
\end{table}

\textbf{Input Size of the Refinement Network}.
The extracted patches are resized into a larger scale before refinement.
Table~\ref{Tab:input_size} shows the impact of input size.
We also report the approximate inference speed of the refinement network, with a fixed batch size of 135 (on average 135 patches per image).
As the input size increases, the AP and AF scores increase accordingly, and slightly drop after 256.
The boundaries can be processed with the higher resolution with the larger input size, thus more details can be retained.

\begin{table}[t]
\begin{center}
  \begin{tabular}{c|c|c|c|ccc}
    size  &FPS &AP &AF &AP$_{S}$ &AP$_{M}$ &AP$_{L}$ \\ \hline
    --    &-- &36.4 &54.9 &11.1 &32.4 &57.3 \\ \hline
    64    &17.5 &39.1 &64.9 &11.8 &35.1 &61.6 \\
    128   &9.4 &39.8 &66.8 &12.7 &\textbf{35.9} &62.2 \\
    256   &4.1 &\textbf{40.0} &\textbf{67.0} &\textbf{12.8} &\textbf{35.9} &\textbf{62.5} \\
    512   &$<$2  &39.7 &66.9 &12.7 &35.7 &61.9 \\
  \end{tabular}
\end{center}
\vspace{-6mm}
  \caption{
    \textbf{Input size of the refinement network}:
    Better performance is achieved with input size of 256$\times$256.
  }
  \label{Tab:input_size}
\end{table}

\textbf{Alternatives of refinement network}.
We compared different backbones for our refinement network in Table~\ref{Tab:segnet}.
As shown, a stronger backbone usually lead to higher performance, but at the expense of lower speed.
Since the model essentially performs binary segmentation for patches, it can further benefit from the advances in semantic segmentation, such as increasing the
resolution of feature maps \cite{chen2017rethinking,chen2018encoder,hrnet} and more effective backbones \cite{xie2017aggregated,zhang2020resnest}.

\begin{table}[t]
\begin{center}
  \begin{tabular}{l|c|c|ccc}
    Net  &FPS &AP &AP$_{S}$ &AP$_{M}$ &AP$_{L}$ \\ \hline
    --         &--  &36.4 &11.1 &32.4 &57.3 \\ \hline
    HRNet-W18s &9.4 &39.8 &12.7 &35.9 &62.2 \\
    HRNet-W18  &5.8 &39.8 &12.6 &35.8 &62.1 \\
    HRNet-W48  &2.5 &40.1 &\textbf{12.9} &36.2 &62.1 \\
  \end{tabular}
\end{center}
  \vspace{-6mm}
  \caption{
    \textbf{Alternatives of the refinement network}:
    Stronger segmentation backbones lead to better results.
  }
  \label{Tab:segnet}
\end{table}

\textbf{NMS Eliminating Threshold}.
We studied the impact of different NMS eliminating thresholds during inference, shown in Table~\ref{Tab:nms_threshold}.
As the threshold gets larger, the number of boundary patches increases rapidly.
The overlap of adjacent patches provides a chance to correct unreliable predictions of the inferior patches.
As shown, the resulting boundary quality was consistently improved with a larger threshold, and reached saturation around 0.55.
We believe a better speed/accuracy trade-off can be achieved by setting a proper threshold.

\begin{table}[t]
\begin{center}
  \begin{tabular}{cc|cc|c}
    thr. &\#patch/img &AP &AP$_{50}$ &AF \\ \hline
    --    &--   &36.4 &60.8 &54.9 \\ \hline
    0    &32  &37.7 &61.5   &58.7 \\
    0.15 &103 &39.6 &61.9   &66.0 \\
    0.25 &135 &39.8 &62.0   &66.8 \\
    0.35 &178 &39.9 &62.0   &67.0 \\
    0.45 &241 &40.0 &62.0   &67.0 \\
    0.55 &332 &\textbf{40.1} &62.0 &67.1 \\
    0.65 &485 &\textbf{40.1} &62.0 &\textbf{67.2} \\
  \end{tabular}
\end{center}
\vspace{-6mm}
  \caption{
    \textbf{NMS eliminating threshold}:
    We achieved consistent gains with the larger thresholds, saturating around 0.55.
    The average number of patches per image is also listed.
  }
  \label{Tab:nms_threshold}
\end{table}

\begin{table*}[ht]
  \small
\begin{center}
  \begin{tabular}{l|l|c|cc|cccccccc}
     &training data &AP$_{\texttt{val}}$ &AP &AP$_{50}$ &person &rider &car &truck &bus &train &mcycle &bicycle \\ \Xhline{1.0pt}
    SGN \cite{SGN} &fine  + coarse &29.2 &25.0 &44.9 &21.8 &20.1 &39.4 &24.8 &33.2 &30.8 &17.7 &12.4 \\
    Mask R-CNN \cite{maskrcnn} &fine &31.5 &26.2 &49.9 &30.5 &23.7 &46.9 &22.8 &32.2 &18.6 &19.1 &16.0 \\
    BMask R-CNN \cite{bmaskrcnn} &fine &35.0 &29.4 &54.7 &34.3 &25.6 &52.6 &24.2 &35.1 &24.5 &21.4 &17.1 \\
    AdaptIS \cite{sofiiuk2019adaptis} &fine &36.3 &32.5 &52.5 &31.4 &29.1 &50.0 &31.6 &41.7 &39.4 &24.7 &12.1 \\
    PANet \cite{panet} &fine &36.5 &31.8 &57.1 &36.8 &30.4 &54.8 &27.0 &36.3 &25.5 &22.6 &20.8 \\
    SSAP \cite{SSAP} &fine &37.3 &32.7 &51.8 &35.4 &25.5 &55.9 &33.2 &43.9 &31.9 &19.5 &16.2 \\
    UPSNet \cite{upsnet} &fine + COCO &37.8 &33.0 &59.7 &35.9 &27.4 &51.9 &31.8 &43.1 &31.4 &23.8 &19.1 \\
    PANet \cite{panet} &fine + COCO &41.4 &36.4 &63.1 &41.5 &33.6 &58.2 &31.8 &45.3 &28.7 &28.2 &24.1 \\ \hline
    Mask R-CNN$^*$ \cite{maskrcnn} &fine + COCO &36.8 &32.6 &59.2 &36.7 &29.2 &52.8 &30.0 &40.3 &27.9 &25.0 &19.0 \\
    + SegFix$^*$ \cite{segfix} & &38.2 &33.3 &57.8 &37.9 &30.3 &54.1 &31.0 &40.0 &27.9 &25.1 &20.5 \\
    + BPR & &41.1 &36.9 &61.0 &42.0 &33.3 &59.9 &32.9 &44.4 &32.6 &28.0 &22.3 \\
    + SegFix + BPR & &40.9 &36.8 &59.8 &41.0 &32.8 &58.7 &32.9 &43.1 &36.8 &26.5 &22.2 \\ \hline
    PolyTransform \cite{polytransform} &fine + COCO &44.6 &40.1 &65.9 &42.4 &34.8 &58.5 &39.8 &50.0 &41.3 &30.9 &23.4 \\
    + SegFix \cite{segfix} & &- &41.2 &66.1 &44.3 &35.9 &60.5 &40.5 &51.2 &41.6 &31.7 &24.1 \\
    + BPR$^\dagger$  & &\textbf{46.9} &42.4 &\textbf{66.6} &45.6 &36.7 &62.4 &41.2 &52.3 &43.4 &\textbf{32.7} &\textbf{25.2} \\
    + SegFix + BPR$^\dagger$ & &- &\textbf{42.7} &66.5 &\textbf{46.0} &\textbf{37.1} &\textbf{62.8} &\textbf{41.3} &\textbf{52.7} &\textbf{43.7} &32.6 &25.1 \\
  \end{tabular}
\end{center}
  \vspace{-5mm}
  \caption{
    \textbf{Results on Cityscapes val (AP$_{\texttt{val}}$ column) and test (remaining columns) sets.}
    We used BPR to denote our framework.
    BPR$^\dagger$ indicates that the BPR trained on the results of Mask R-CNN$^*$ was transferred to another model.
    Mask R-CNN$^*$ is on our implementation, slightly higher than \cite{maskrcnn}.
    SegFix$^*$ used their own Mask R-CNN baseline ($36.5/32.0$ in AP val/test), slightly lower than ours.
    We established the new state-of-the-art results on Cityscapes \texttt{val} and \texttt{test} sets.
  }
  \label{Tab:cityscapes_sota}
\end{table*}

\subsection{Transferability}
What the BPR model learned is a general ability to correct error pixels around instance boundaries.
We can easily transfer this \textbf{ability of boundary refinement} to refine the results of any instance segmentation model.
Specifically, once we get a model trained on the boundary patches extracted from the train-set predictions of Mask R-CNN on Cityscapes,
we can make inference to refine any predictions (on Cityscapes train/val/test sets) produced by any models (not only Mask R-CNN), without the need for training from scratch.
After training, the BPR model becomes model-agnostic, similar to SegFix \cite{segfix}.
We validated the transferability by applying the model trained on Mask R-CNN results to refine the predictions of PointRend \cite{pointrend} and SegFix \cite{segfix}.
Note that these two methods are also designed to improve boundary quality in segmentation.
As shown in Table~\ref{Tab:transfer}, the transferred model still improved the results of PointRend and SegFix by a large margin, suggesting that our method is compatible with them.

\begin{table}[t]
\begin{center}
  \begin{tabular}{c|cc|c}
      &AP &AP$_{50}$ &AF \\ \hline
    PointRend \cite{pointrend}   &35.6 &60.6 &58.0 \\
    w/ BPR$^\dagger$     &38.6 &62.4 &66.5 \\ \hline
    Mask R-CNN + SegFix \cite{segfix}     &38.2 &63.4 &63.2 \\
    w/ BPR$^\dagger$     &40.0 &63.4 &67.0 \\
  \end{tabular}
\end{center}
\vspace{-5mm}
  \caption{
    \textbf{Transfer to Other Models}:
    BPR$^\dagger$ was trained on the results of Mask R-CNN.
    It can be successfully transferred to refine the results of PointRend and SegFix.
  }
  \label{Tab:transfer}
\end{table}

\begin{table}[t]
\begin{center}
  \begin{tabular}{c|cc|ccc|c}
    w/ BPR  &AP &AP$^\star$ &AP$_{S}^\star$ &AP$_{M}^\star$ &AP$_{L}^\star$ &AF \\ \hline
           &38.4 &40.4 &24.5 &48.3 &57.2 &54.5 \\
    \cmark &\textbf{39.2} &\textbf{42.1} &\textbf{24.8} &\textbf{50.3} &\textbf{60.4} &\textbf{58.4} \\
  \end{tabular}
\end{center}
\vspace{-5mm}
  \caption{
    \textbf{Results on COCO}.
    AP$^\star$ is measured on the higher-quality LVIS \cite{lvis} annotations.
    We improved based on the results of Mask R-CNN ResNeXt-FPN-101 baseline.
  }
  \label{Tab:coco}
\end{table}

\subsection{Overall Results}

\textbf{Comparison with State-of-the-art Methods}.
We integrated the optimal design choices and hyperparameters found in above ablation experiments into a stronger BPR model.
Specifically, we adopted the HRNetV2-W48 as our refinement network, with 256$\times$256 input patches resized from 64$\times$64, and a NMS threshold of 0.55 during inference.
We evaluated the framework on Cityscapes \texttt{val} and \texttt{test} sets and compared the performance against some state-of-the-art methods in Table~\ref{Tab:cityscapes_sota}.
(1) Compared with the Mask R-CNN baseline, we achieved a significant improvement (+$\mathbf{4.3\%}$ AP in both \texttt{val} and \texttt{test}).
We outperformed SegFix \cite{segfix} by a large margin, which is also a boundary refinement module applied to the same baseline with ours.
Furthermore, by applying our BPR model to the results already refined by SegFix, we can still improved a lot (slightly lower than applying BPR only).
(2) We transferred the above BPR model to refine the results of the stronger PolyTransform \cite{polytransform} baseline ($1^{st}$ place at CVPR 2020).
Our ``PolyTransform + BPR'' consistently improved $2.3\%$ AP on Cityscapes \texttt{test} set and also outperformed ``PolyTransform + SegFix'' ($2^{nd}$ place at ECCV 2020) by a large margin (+$1.2\%$).
By applying BPR to ``PolyTransform + SegFix'', we established a new state-of-the-art on Cityscapes \texttt{test} with AP of \textbf{42.7\%}, reaching $\mathbf{1^{st}}$ place on the Cityscapes leaderboard by the CVPR 2021 submission deadline.

\begin{figure*}[ht]
\begin{center}
  \includegraphics[width=0.95\linewidth]{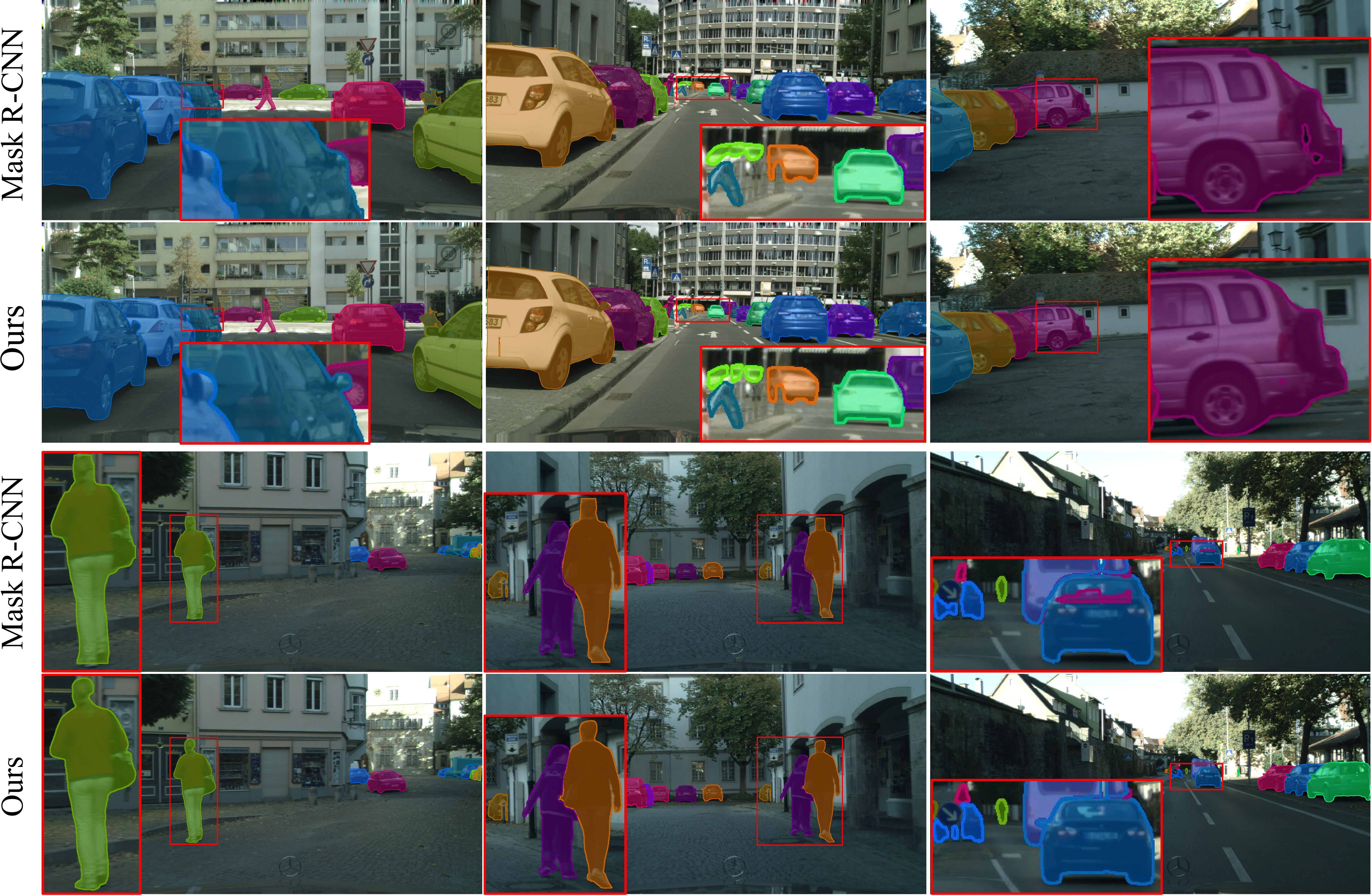}
\end{center}
  \vspace{-5mm}
  \caption{
    Qualitative results on Cityscapes \texttt{val}.
    The proposed framework (2nd and 4th rows) produces substantially better masks with more precise boundaries than Mask R-CNN (1st and 3rd rows).
    Best viewed digitally and in colour.
  }
  \label{fig:visualization}
\end{figure*}

\textbf{Qualitative Results}.
We show some qualitative results on Cityscapes \texttt{val} in Figure~\ref{fig:visualization}.
Compared with the coarse predictions of Mask R-CNN, our BPR framework generated substantially better instance segmentation results with precise and distinct boundaries.
It largely alleviated the over-smoothing issues \cite{pointrend} in previous methods caused by the low resolution feature maps.
More results are included in Appendix \ref{appendix_e}.
In addition, we also provided a detailed limitation analysis in Appendix \ref{appendix_f}.

\textbf{Speed}.
Only the speed of refinement network was considered in Table~\ref{Tab:input_size} and \ref{Tab:segnet}, excluding the patch extraction and reassembling time.
As a whole pipeline, it takes about 211ms to process a single Cityscapes image (1024$\times$2048) on a single RTX 2080Ti GPU under the default setting of ablation experiments, which is still much faster than PolyTransform \cite{polytransform}.
The detailed speed calculation and more speed analysis are included in Appendix \ref{appendix_c}.

\textbf{Results on COCO Dataset}.
To demonstrate the generality of our framework, we also report the results on the more challenging COCO dataset \cite{mscoco}, which contains 80 categories and more images (118k/5k for train/val).
It is important to note that the coarse annotations in COCO may not fully reflect the improvements in mask quality \cite{lvis}.
Following PointRend \cite{pointrend}, we further report the AP$^\star$ measured using the higher quality LVIS \cite{lvis} annotations.
We randomly sampled about $8\%$ of instances for fast training.
As shown in Table~\ref{Tab:coco}, we improved the powerful Mask R-CNN ResNeXt-FPN-101 baseline by $0.8\%$ AP and $1.7\%$ AP$^\star$ on \texttt{val2017}.
The coarse annotations on COCO \texttt{train2017} may provide ambiguous optimization objectives, especially for our local boundary patches. It may mislead the learning of our BPR model, leading to suboptimal results.
This issue was also observed in some contour-based instance segmentation methods \cite{deepsnake,xie2020polarmask,xu2019explicit}.
We believe that training with more instances on higher quality annotations (\eg LVIS) can further improve the results.
More analysis on COCO dataset is included in Appendix \ref{appendix_d}.

\section{Conclusion}
In this paper, we propose a conceptually simple yet effective boundary refinement framework to improve the boundary quality for any instance segmentation model.
Starting from a coarse instance mask, we extract and refine a series of boundary patches along the predicted instance boundaries through an effective refinement network.
The proposed framework achieved consistent and impressive improvements based on different baselines.
Qualitative results show that our approach produced high-quality masks with precise and distinct boundaries.

\noindent\textbf{Acknowledgements} \
This work was supported by the National Key Research and Development Program of China (No. 2017YFA0700904), the National Science and Technology Major Project (No. 2018ZX01028-102), the National Natural Science Foundation of China (Nos. 61836014, U19B2034, 62061136001 and 61620106010) and THU-Bosch JCML center.

{\small
\bibliographystyle{ieee_fullname}
\bibliography{egbib}
}

\appendix

\section*{Appendix}

\setcounter{table}{0}
\setcounter{figure}{0}
\setcounter{equation}{0}
\renewcommand{\theequation}{S\arabic{equation}}
\renewcommand{\thefigure}{S\arabic{figure}}
\renewcommand{\thetable}{S\arabic{table}}
\renewcommand{\theHfigure}{S\arabic{figure}}
\renewcommand{\theHtable}{S\arabic{table}}

\section{Implementation Details} \label{appendix_a}
We adopted the \texttt{MMSegmentation} \cite{mmsegmentation} codebase to implement the boundary patch refinement network.
We almost followed the same training protocol as HRNet.
The image patches are augmented by random horizontal flipping and random photometric distortion.
The binary mask patches are normalized with the mean and standard deviation both equal to 0.5.
We use the SGD optimizer with the initial learning rate of 0.01, the momentum of 0.9, and the weight decay of 0.0005.
The learning rate is decayed using the poly learning rate policy with the power of 0.9.
The models are trained for 160K iterations with a batch size of 32 on 4 GPUs.
Taking the default setting adopted in ablation studies as example, we extracted 280k/67k patches from the train/val results of Mask R-CNN (adopted from \texttt{MMDetection} \cite{mmdetection}).
It takes about 10 hours of training on 4 NVIDIA RTX 2080Ti GPUs under this setting.

\section{Different Patch Extraction Schemes} \label{appendix_b}
In Section \ref{ablation_study}, we compared the proposed ``dense sampling + NMS'' scheme with another two patch extraction schemes: \textit{pre-defined grid} and \textit{instance-level patch}.
Here we provide the implementation details and further analysis of these two schemes.
As illustrated in Figure~\ref{fig:extraction_schemes_b}, the \textit{pre-defined grid} scheme simply divides the input image into a group of patch candidates according to a pre-defined grid.
Candidates that covering both foreground and background pixels are choosen as boundary patches for refinement.
This straightforward scheme yields plenty of inferior patches, as indicated by yellow dashed boxes in Figure~\ref{fig:extraction_schemes_b},
which have the imbalanced foreground/background ratio and may lack of real boundary cues, thus leading to sub-optimal results.
Another scheme is extracting the \textit{instance-level patch} (Figure~\ref{fig:extraction_schemes_c}) based on the detected bounding box, which is similar to previous studies \cite{polytransform,liu20201st}.
This scheme can be viewed as an improved Mask R-CNN equipped with a stand-alone mask head, while still fails to solve the optimization bias issue and the learning process is dominated by interior pixels.
Different from these methods, by adaptively extracting patches along the predicted boundaries in a \textit{sliding-window} style (Figure~\ref{fig:extraction_schemes_a}) and refining the local boundary regions separately, the above issues can be alleviated.

\section{More Speed Analysis} \label{appendix_c}
The inference time of our proposed framework is independent of the original instance segmentation models,
which consists of three parts: patch extraction, refinement, and reassembling.
Note that only the refinement part was considered when calculating the FPS in Table \ref{Tab:input_size} and \ref{Tab:segnet}.
Besides, the FPS was measured in an imprecise manner by fixing the batch size to 135 (average number of patches per image), while the exact number of patches varies from image to image.
Here we report the total inference time, which measured by calculating the exact inference time for each image individually and then averaging them.
Taking the default setting (HRNet-W18s with input size of 128$\times$128) in our ablation experiments as example, it takes about 211ms (52ms,81ms,78ms for the above three parts respectively) to process an image (1024$\times$2048) of Cityscapes
on a single RTX 2080Ti GPU, which is still much faster than PolyTransform (575ms\footnote{Measured on a single GTX 1080Ti GPU, which is about $35\%$ slower than our RTX 2080Ti GPU with FP32 training (ref. \url{lambdalabs.com}).} per image \cite{polytransform}).
Undoubtedly, the network speed can be further improved with more efficient backbones (\eg MobileNets), smaller input size (\eg 32$\times$32 or 64$\times$64), and less inference patches (\eg with lower NMS thresholds or adaptively selecting the most unreliable patches).
Note that the BPR models can still achieve a remarkable performance under these lightweight settings (Tables \ref{Tab:input_size},\ref{Tab:segnet},\ref{Tab:nms_threshold}).
The patch extraction and reassembling steps can also be accelerated with more CPU cores.

\section{More Analysis on COCO Dataset} \label{appendix_d}
\begin{figure*}[t]
\begin{center}
  \includegraphics[width=1.0\linewidth]{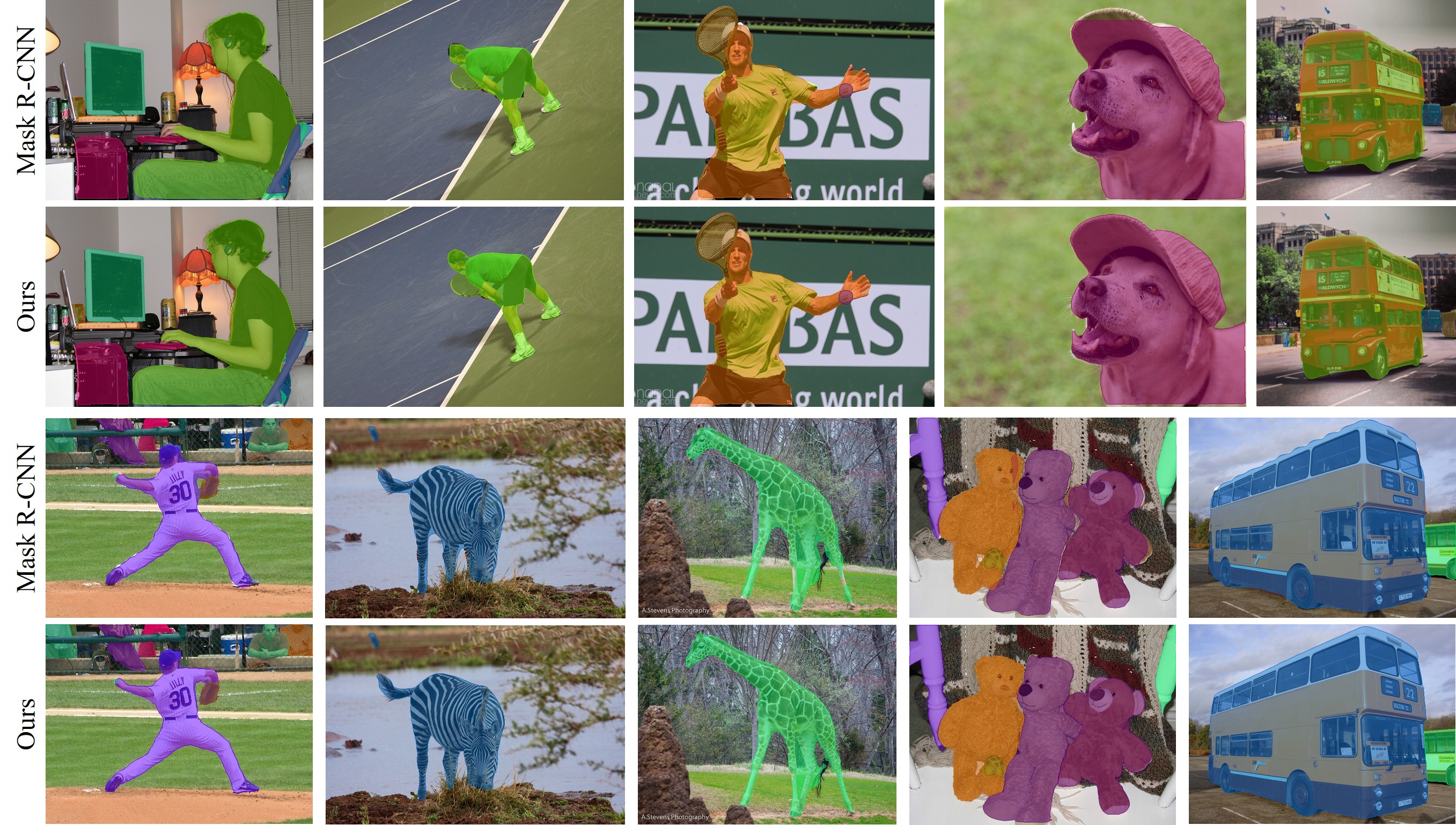}
\end{center}
  \vspace{-2mm}
  \caption{
    Qualitative results on COCO \texttt{val2017}.
    The proposed framework (2nd and 4th rows) generates substantially better masks with more precise boundaries than Mask R-CNN (1st and 3rd rows).
    Best viewed digitally and in colour.
  }
  \label{fig:coco_pred}
\end{figure*}

In theory, the proposed framework, as a general boundary refinement mechanism, can be applied to any instance segmentation dataset.
We achieved impressive performance on Cityscapes, while the AP improvement on COCO dataset was not as high as we got on Cityscapes (see Table \ref{Tab:coco}).
The most critical problem is that the coarse polygon-based annotations on COCO dataset yield significantly lower boundary quality \cite{lvis}.
Several examples (which are ubiquitous on COCO) are shown in Figure~\ref{fig:coco_gt}.
The misalignment between annotations and real instance boundaries may greatly increase the optimization difficulty of our refinement model.
Especially, the coarse annotations may provide ambiguous optimization objectives for our local boundary patches, thus hampering the model convergence.
We observed that some \textit{contour-based} instance segmentation methods \cite{deepsnake,xie2020polarmask,xu2019explicit}, which are sensitive to the quality of boundary annotations, also suffered from this misalignment issue.
It seems that the coarse COCO annotations may not friendly to these methods and it is hard to achieve very high AP scores based on these approaches.
In spite of this, we still significantly improved the Mask R-CNN results in some cases, shown in Figure~\ref{fig:coco_pred}.
Some results are even better than their annotations (the first three examples in Figures~\ref{fig:coco_gt}, \ref{fig:coco_pred}).

\begin{figure}
     \centering
     \begin{subfigure}[b]{0.45\textwidth}
         \centering
         \includegraphics[width=\textwidth]{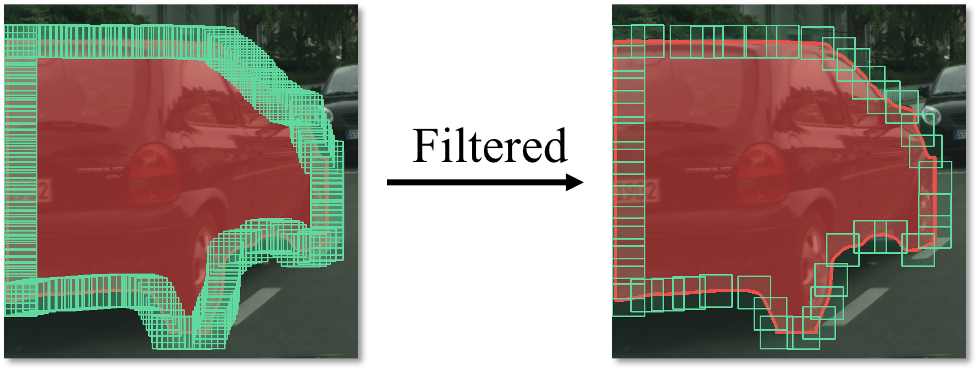}
         \vspace{-5mm}
         \caption{dense sampling + NMS filtering}
         \label{fig:extraction_schemes_a}
     \end{subfigure}
     \hfill
     \begin{subfigure}[b]{0.45\textwidth}
         \centering
         \includegraphics[width=\textwidth]{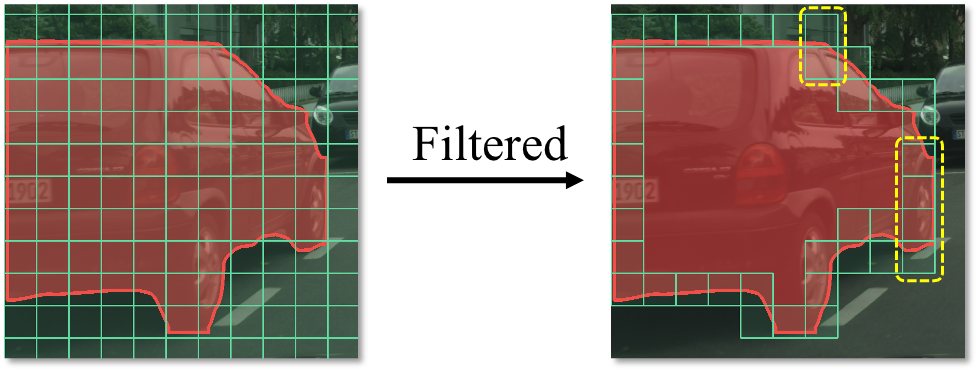}
         \vspace{-5mm}
         \caption{pre-defined grid}
         \label{fig:extraction_schemes_b}
     \end{subfigure}
     \hfill
     \begin{subfigure}[b]{0.45\textwidth}
         \centering
         \includegraphics[width=\textwidth]{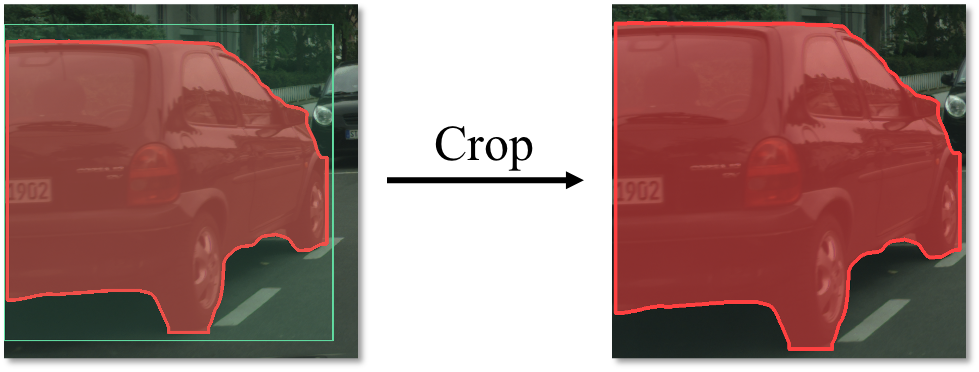}
         \vspace{-5mm}
         \caption{instance-level patch}
         \label{fig:extraction_schemes_c}
     \end{subfigure}
        \vspace{-2mm}
        \caption{
          Illustration of three different patch extraction schemes.
          Best viewed digitally and in colour.
        }
        \label{fig:extraction_schemes}
\end{figure}

\begin{figure}[t]
  \setcounter{figure}{5}
\begin{center}
  \includegraphics[width=0.8\linewidth]{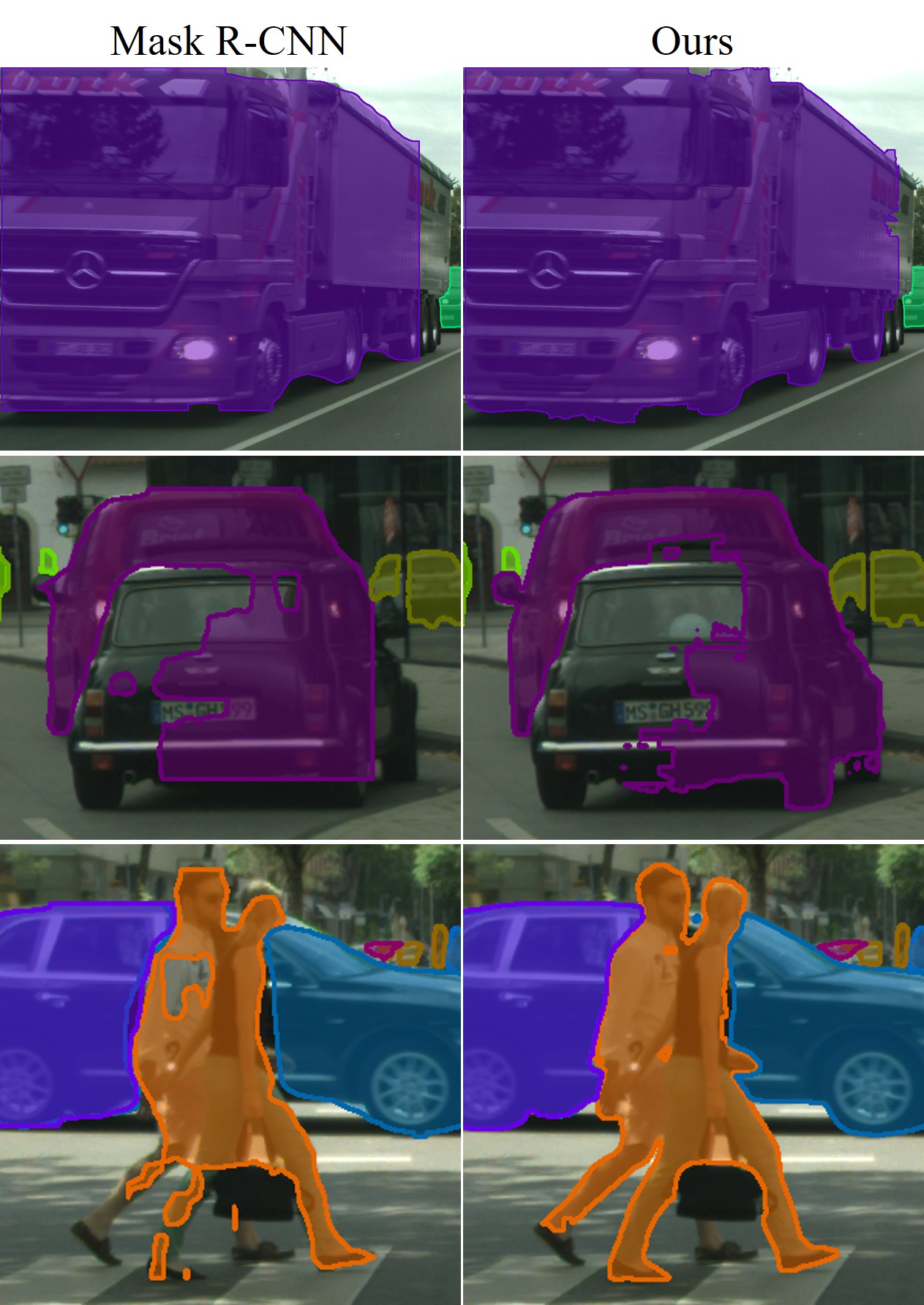}
\end{center}
\vspace{-5mm}
  \caption{
    Illustration of some failure cases.
  }
  \label{fig:supp_failure}
\end{figure}

\begin{figure}[t]
\begin{center}
  \includegraphics[width=1.0\linewidth]{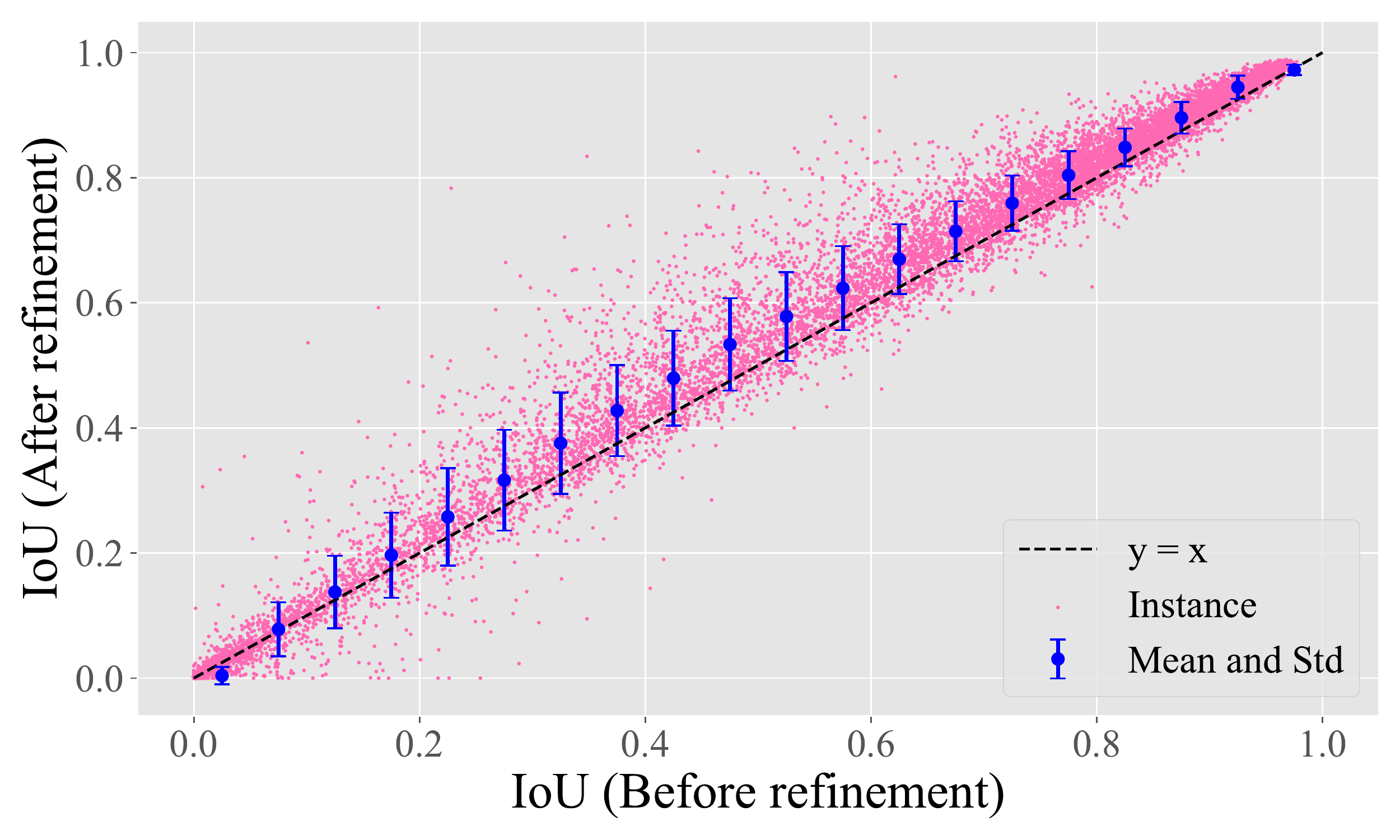}
\end{center}
\vspace{-6mm}
  \caption{
    IoU improvements for all predicted instances on Cityscapes val set.
    Each red dot indicates an instance.
    Dots below the dash line are failure cases.
  }
  \label{fig:iou_curve}
\end{figure}

\section{More Qualitative Results} \label{appendix_e}
\begin{figure*}[ht]
\begin{center}
  \includegraphics[width=0.94\linewidth]{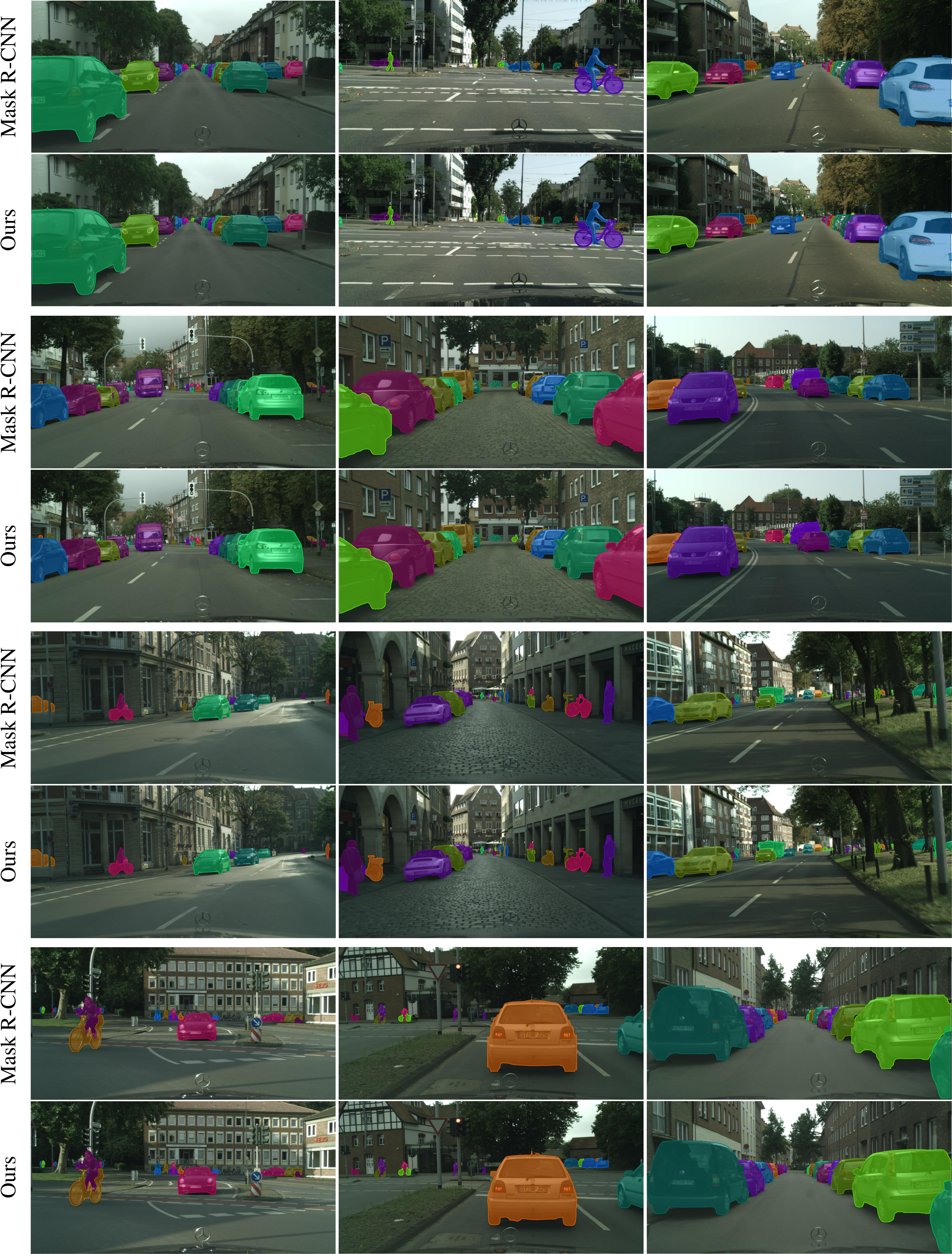}
\end{center}
  \vspace{-5mm}
  \caption{
    Qualitative results on Cityscapes \texttt{val}.
    The proposed framework (2nd and 4th rows) produces substantially better masks with more precise boundaries than Mask R-CNN (1st and 3rd rows).
    Best viewed digitally and in colour.
  }
  \label{fig:supp_vis_image}
\end{figure*}

\begin{figure*}[ht]
\begin{center}
  \includegraphics[width=0.95\linewidth]{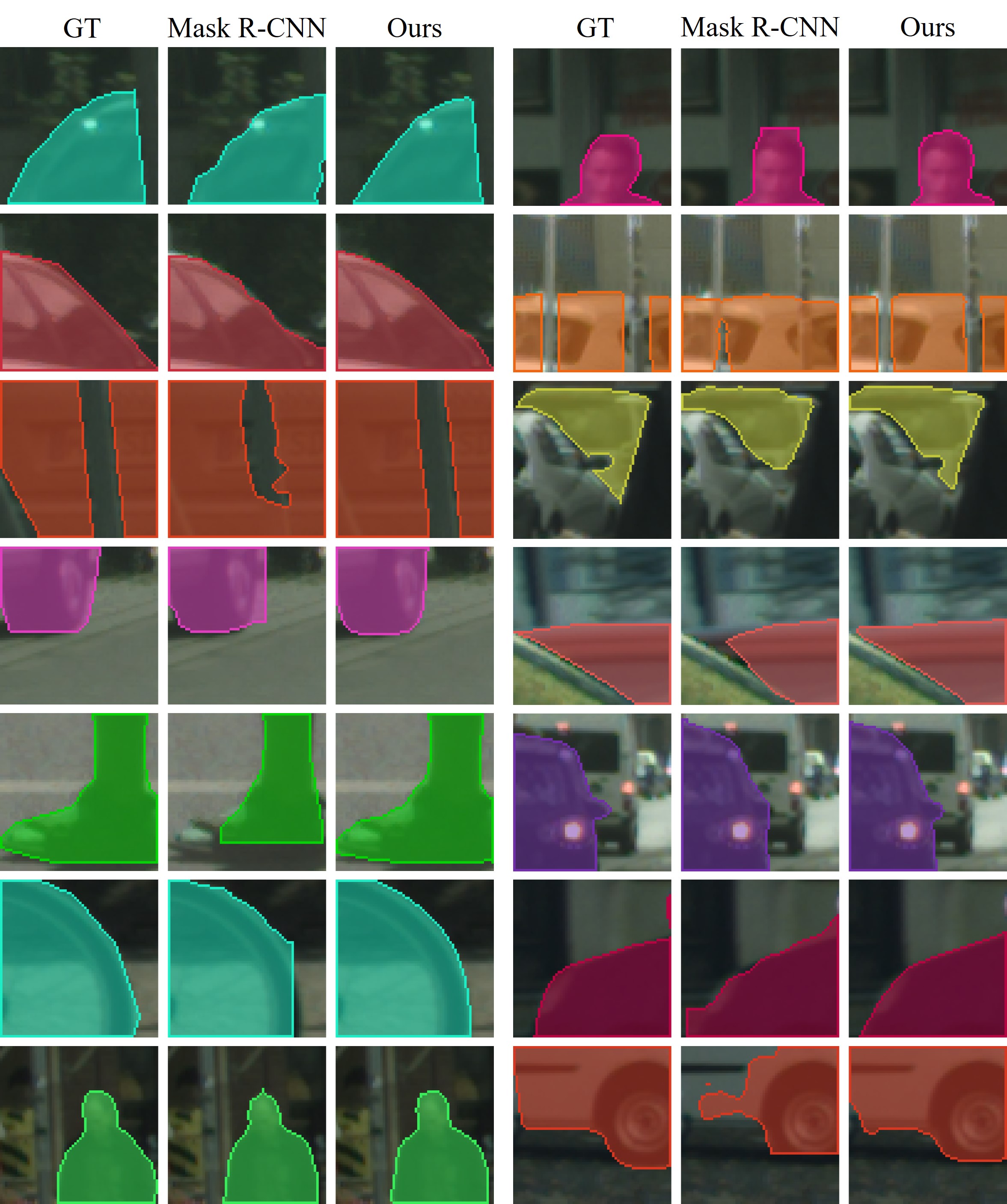}
\end{center}
  \vspace{-2mm}
  \caption{
    Boundary patch examples of: ground-truth (1st and 4th columns), predictions of Mask R-CNN (2nd and 5th columns), results refined by our proposed framework (3rd and 6th columns).
    Best viewed digitally and in colour.
  }
  \label{fig:supp_vis_patch}
\end{figure*}

We provid more qualitative results on Cityscapes \texttt{val}, including \textit{image-level} (Figure~\ref{fig:supp_vis_image}) and \textit{patch-level} (Figure~\ref{fig:supp_vis_patch}) results.
As shown, our proposed framework consistently improves the instance segmentation results of Mask R-CNN and produces substantially better instance masks with more precise boundaries.

\section{Limitation Analysis} \label{appendix_f}
The performance of our proposed framework relies on the boundary quality of initial masks.
Some failure cases are illustrated in Figure~\ref{fig:supp_failure}.
For example, our model failed to produce an optimal mask if the initially predicted boundaries are far from the real object boundaries (1st row), but note that we still refined this case to some extent (IoU was improved).
In addition, if the initial mask largely over-segments the neighboring instance, our model may regard the two instances as a whole and further enlarge this error (2nd and 3rd rows) since we only process the local boundary regions without a global view.
We analyzed the IoU improvements for all predicted instances on Cityscapes val set, shown in Figure~\ref{fig:iou_curve}.
In most cases, our refinement model can effectively improve the mask IoU (red dots above the dash line).
However, we found that it's hard to refine instance masks with extremely lower IoU (\eg $<$ 0.1) due to the poor quality of initial boundaries.
In addition, we observed that the improvement for smaller instances (about $2\%$ in AP$_{S}$) is not as high as we got for larger instances (about $5\%$ in AP$_{L}$).
Compared to the upper-bound results (see Table \ref{Tab:intro_gt_improves}), there is still a large step to take for boundary refinement, especially for small instances.

\section{More Transferring Results} \label{appendix_g}
In Table \ref{Tab:transfer}, we verified that the BPR model trained on Mask R-CNN results can be effectively transferred to refine the results of PointRend and SegFix.
As an opposite directions with Table \ref{Tab:transfer}, we instead trained the BPR model on PointRend or SegFix results and transferred them to refine the Mask R-CNN predictions.
As shown in Table \ref{Tab:supp_transfer}, the transferring is also workable.

\begin{table}[t]
\small
\begin{center}
  \begin{tabular}{l|cc|c}
      &AP &AP$_{50}$ &AF \\ \hline
    Mask R-CNN   &36.4 &60.8 &54.9 \\ \hline
    w/ BPR (trained on PointRend \cite{pointrend})  &38.7 &61.5 &64.2 \\
    w/ BPR (trained on SegFix \cite{segfix})     &39.1 &61.7 &64.1 \\
  \end{tabular}
\end{center}
\vspace{-5mm}
  \caption{
    BPR models were trained on PointRend and SegFix results respectively, and transferred to refine the Mask R-CNN predictions.
  }
  \label{Tab:supp_transfer}
\end{table}


\begin{figure*}[t]
  \setcounter{figure}{1}
\begin{center}
  \includegraphics[width=1.0\linewidth]{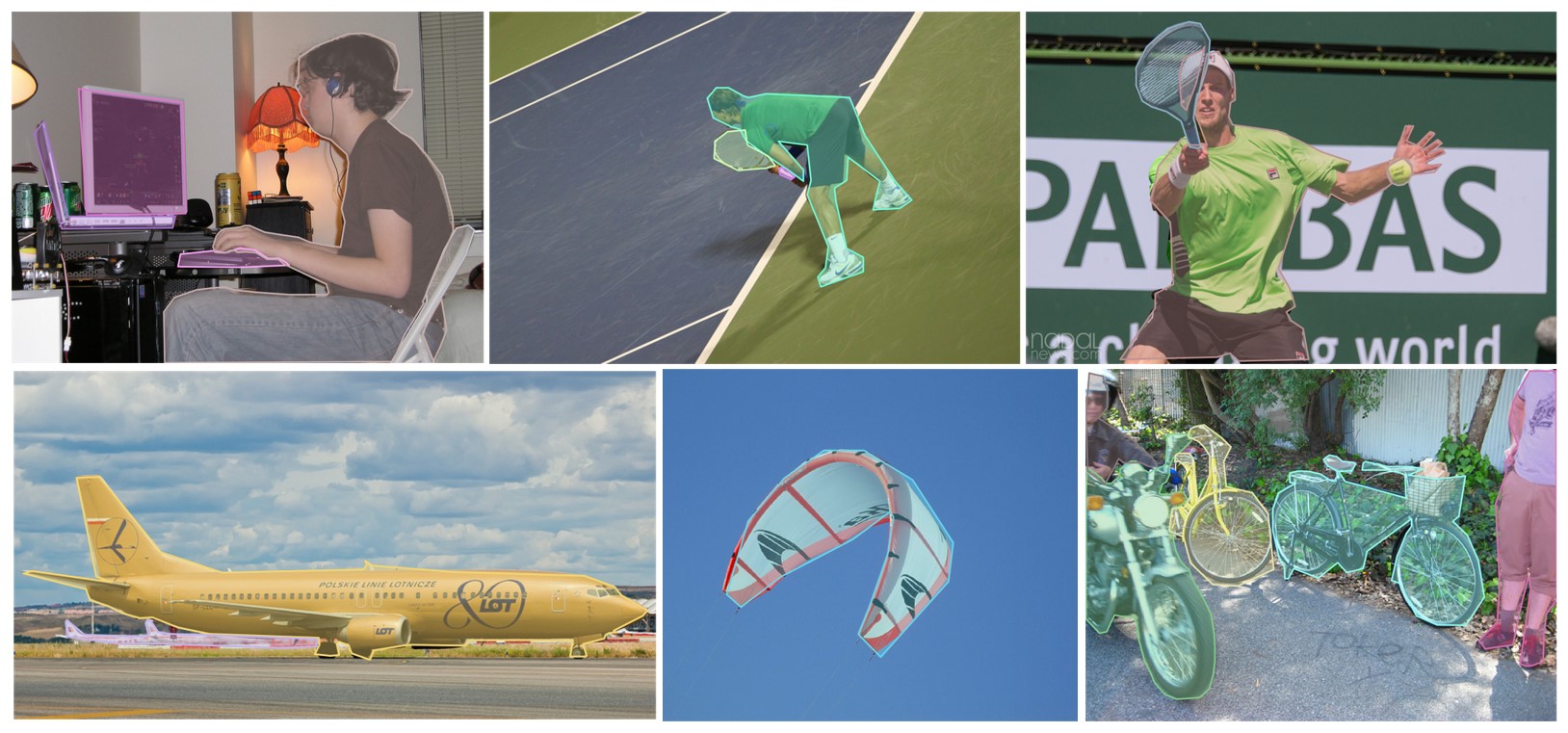}
\end{center}
  \vspace{-3mm}
  \caption{
    Illustration of the coarse annotations on COCO \texttt{val2017}.
    The annotated instance masks are not well-aligned with the real object boundaries.
    Best viewed digitally and in colour.
  }
  \label{fig:coco_gt}
\end{figure*}

\end{document}